\pdfoutput=1

\documentclass[11pt]{article}

\usepackage[preprint]{acl}

\usepackage{times}
\usepackage{latexsym}
\usepackage{enumitem}
\usepackage[T1]{fontenc}

\usepackage[utf8]{inputenc}

\usepackage{microtype}

\usepackage{inconsolata}

\usepackage{graphicx}

\usepackage{amsmath}
\usepackage{bbm}
\usepackage{subcaption}

\usepackage{pgfplots}
\usepackage{xcolor}
\pgfplotsset{compat=1.9}

\title{Is This Collection Worth My LLM's Time? Automatically Measuring Information Potential in Text Corpora}

\author{
  Tristan Karch\thanks{Equal contribution.} \\
  EPFL / DHLab \\
  \texttt{tristan.karch@gmail.com} 
  \And
  Luca Engel$^*$ \\
  EPFL / DHLab \\
  \\
  \\\AND
  Philippe Schwaller \\
  EPFL / ILIAC \\
  \And 
  Frédéric Kaplan \\
  EPFL / DHLab \\
}

\begin{document}
\maketitle
\begin{abstract}
As large language models (LLMs) converge in capabilities, progress increasingly depends on integrating novel, high-value information sources. Yet deciding which text collections justify the costs of digitization and integration remains difficult. We introduce an automated pipeline that estimates a collection’s information potential without retraining. The method generates multiple-choice questions (MCQs) from texts and compares model performance with and without access to the source; the gap serves as a proxy for novel knowledge. We validate this approach on five datasets—EPFL PhD manuscripts, Venetian historical records, two Wikipedia subsets, and a synthetic baseline. Results show that the pipeline reliably distinguishes high-value from redundant corpora, offering a practical tool for prioritizing data acquisition and integration.

\end{abstract}

\section{Introduction}

Large language models (LLMs) show convergent capabilities as they scale, largely independent of architecture \cite{huh2024platonicrepresentationhypothesis}. Combined with scaling laws \cite{kaplan2020scalinglawsneurallanguage} and the shift toward data-centric AI \cite{zha2025datacentric}, this suggests that future progress depends less on model design and more on integrating high-quality, novel information sources. Yet acquiring such data is costly: digitization, preprocessing, and fine-tuning require substantial resources.

This raises a central question: how can we assess whether a text collection meaningfully expands an LLM’s knowledge before investing in its integration? Existing strategies typically require expensive retraining or fine-tuning, while retrieval-augmented generation (RAG) \cite{lewis2020rag} mitigates some costs but still requires careful knowledge curation \cite{pmlr-v202-kandpal23a}. Practical methods for estimating the value of new corpora remain limited.

We propose an automated pipeline that estimates a collection’s information potential without training. The method generates MCQs from text chunks, applies a two-stage quality filter---(i) context–answer alignment using ROUGE-L/Jaccard and (ii) distractor plausibility via embedding cosine similarity---and then compares LLM performance with and without access to the source. The performance gap serves as a proxy for novel information.

To validate this approach, we test three distinct datasets: (1) EPFL PhD manuscripts, containing specialized academic knowledge, (2) Wikipedia articles, representing widely available content, and (3) a synthetic baseline of model-generated text. Importantly, the pipeline is dataset-agnostic and can be applied to any domain, making it a practical tool for prioritizing corpora for fine-tuning or RAG systems.

\section{Related Work}

\noindent\textbf{Information Gain. }
The challenge of efficiently selecting new information sources for LLMs can be viewed through the lens of optimal experiment design, a field pioneered by \citet{fedorov1972theory}. This framework emphasizes maximizing information gain while being strategic about resource allocation -- a particularly relevant consideration given the costs associated with integrating new data into LLM systems. Information gain itself has been conceptualized across various fields: in information theory, it relates to reductions in algorithmic information content \cite{cover1989kolmogorov}; in machine learning, it quantifies a feature's contribution to model performance \cite{ODHIAMBOOMUYA2021114765}; and in cognitive science, it represents uncertainty reduction in our experience of the world \cite{damiano2021visual}. While these theoretical frameworks provide valuable insights, they have not been previously applied to the specific challenge of evaluating the potential value of text collections for enhancing LLM knowledge. Our work bridges this gap by proposing a practical, MCQ-based approach that quantifies information gain by measuring an LLM's ability to answer questions about a text collection with and without access to the source material.\\

\noindent\textbf{Knowledge Detection in LLMs. }
Prior research has developed several methods to analyze how LLMs process and retain textual information. Work on memorization \cite{hartmann2023sokmemorizationgeneralpurposelarge,shi2024detectingpretrainingdatalarge} and data contamination \cite{yax2024assessingcontaminationlargelanguage,golchin2024datacontaminationquiztool} focuses on identifying verbatim recall of training data, while hallucination detection \cite{farquhar_detecting_hallucinations_llms_semantic_entropy} aims to identify when models generate false information. Research on novelty detection has primarily focused on linguistic and semantic novelty \cite{language_novelty_1,semantic_linguistic_novelty_focus_paper}, with less attention paid to factual novelty. While these approaches provide valuable insights into model behavior, they are retrospective -- analyzing what models have already learned or memorized. In contrast, our work takes a prospective approach, developing metrics to evaluate the potential value of new information sources before investing in their integration into LLM systems.

\section{Methods}

Multiple Choice Questions (MCQs) are a well-established tool for knowledge assessment, supported by research in cognitive science and educational psychology~\cite{Haladyna01072002}. Their four-option format, consisting of one correct answer and three distractors, offers an optimal balance between assessment reliability and cognitive load~\cite{Vyas2008}. MCQs are particularly valuable for automated evaluation as they provide objective correctness measures while efficiently testing understanding across diverse topics~\cite{Oc21102024}. For LLM evaluation, they offer the added advantage of constraining the output space to a finite set of options, eliminating ambiguity in grading and enabling precise comparisons across conditions.

\subsection{MCQ Generation}
The input text is first divided into manageable chunks of 2000 words to ensure consistent context length across questions. These chunks serve as the basis for LLM-generated MCQs, where each question is crafted to test understanding of specific information within the chunk. The exact prompts used for generation can be found in table~\ref{tab:prompts} of the Supplementary Methods.\\

\subsection{MCQ Quality Filtering (Two-Stage)}
To avoid trivial or ambiguous items, we apply complementary filters that test (a) grounding in the source and (b) distractor plausibility:

\paragraph{(a) Context–Answer Alignment.}
For each MCQ, we require the correct answer \(g\) to be better aligned with its source chunk \(c\) than any distractor \(d_i\).
We compute surface-level similarity via Jaccard and ROUGE-L~\cite{lin-2004-rouge}:
\begin{equation}
    \min_i \left[\text{sim}(c, g) - \text{sim}(c, d_i)\right] \quad \text{,} \ i = 1, 2, 3
    \label{eq:similarity_comparison_input_correct_minus_input_distractor}
\end{equation}
where \(\text{sim}\in\{\text{Jaccard},\text{ROUGE-L}\}\).
Items failing either metric are discarded, removing questions whose “correct” answer is not actually supported by the text, or whose distractors inadvertently match the context.

\paragraph{(b) Distractor Plausibility.}
To prevent elimination-by-obviousness while keeping answers unambiguous, we enforce semantic closeness (but not identity) between \(g\) and at least one distractor using cosine similarity over sentence embeddings using the state-of-the-art NV-Embed-v2~\cite{nvidia_embed_v2_first,nvidia_embed_v2_second}:
\begin{equation}
    \max_i \left[ \text{cos-sim}(g, d_i) \right] \quad \text{,} \ i = 1, 2, 3
    \label{eq:similarity_comparison_correct_answer_distractor}
\end{equation}
which retains items where distractors are “nearby” enough to be plausible but not paraphrases of the correct answer. Thresholds are set by percentiles (e.g., 40–60th) computed on the pool of generated items, trading coverage for quality.

\subsection{Evaluation (Position-Debiased, With/Without Context)}
Each filtered MCQ is evaluated in two conditions:
\begin{enumerate}[noitemsep]
\item \textbf{Direct (no context):} The model answers \((q,\{A,B,C,D\})\) without access to the chunk.
\item \textbf{With context:} The model receives the chunk \(c\) plus the same \((q,\{A,B,C,D\})\).
\end{enumerate}
To mitigate position bias, we evaluate each item four times, rotating the correct option through A, B, C, and D while shuffling distractors. A question is marked correct in a condition only if the model selects the correct option for all four rotations (a conservative criterion that reduces spurious gains).

We define the information potential (\(IP\)) of a collection as the normalized improvement in correctness when adding context:
\begin{equation}
IP=\frac{C_{\text{context}}-C_{\text{direct}}}{|\mathcal{Q}|-(I_{\text{context}}+I_{\text{direct}})},
\label{eq:knowledge_gain_formula}
\end{equation}
where \(C_{\text{context}}\) and \(C_{\text{direct}}\) count questions answered correctly (with the “4x” position control) with and without context, and \(I_{\text{context}}, I_{\text{direct}}\) count incorrect ones. The denominator excludes items missed in both settings to focus on questions that are answerable in at least one condition.

\subsection{Data}
We evaluate five collections with distinct expected novelty profiles:
\begin{itemize}[noitemsep]
\item \textbf{EPFL PhD manuscripts:} 177 theses with specialized, recent research likely underrepresented in pretraining corpora.
\item \textbf{Venetian historical records:} Public-domain books on Venice’s urban history, many newly digitized and rare online.
\item \textbf{Wikipedia (EPFL-related) and Wikipedia (Venice-related):} Articles fetched via the Wikipedia API using manuscript titles and Venetian themes as seeds; expected to overlap with pretraining data.
\item \textbf{Synthetic baseline:} Model-generated chunks on the same themes to establish a lower-bound novelty reference.
\end{itemize}
All corpora are chunked uniformly; MCQs, filtering metadata, and prompts are released with this submission to support reproducibility.


\section{Results}

\begin{figure}[h] 
    \centering

    \begin{tikzpicture}
        \begin{axis}[
            width=0.48\textwidth, height=7.5cm,
            xlabel={Cutoff Percentile Values},
            ylabel={Fraction of Correct Responses},
            xmin=0, xmax=60,
            ymin=0.25, ymax=1,
            xtick={0,10,20,30,40,50,60},
            ytick={0.3,0.4,0.5,0.6,0.7,0.8,0.9,1},
            legend pos=south west,
            ymajorgrids=true,
            grid style=dashed,
            legend style={font=\small, 
                          fill opacity=0.7,
                          legend cell align=left,
                          fill=white, 
                          text opacity=1},
        ]

        \addplot[thick, color=cyan] 
        coordinates {(0, 0)};
        \addlegendentry{Cosine}

        \addplot[thick, color=orange] 
        coordinates {(0, 0)};
        \addlegendentry{Rouge-L and Jaccard}

        \addplot[mark=square*, only marks, thick, color=black, mark options={solid, fill=black}] 
            coordinates {(0,0.948)}; 
        \addlegendentry{WC 4x}
        
        \addplot[mark=*, only marks, thick, color=black, mark options={solid, fill=black}] 
            coordinates {(0,0.642)}; 
        \addlegendentry{NC 4x}
        
        \addplot[mark=triangle*, only marks, thick, color=black, mark options={solid, fill=black}] 
            coordinates {(0,1)}; 
        \addlegendentry{Questions Remaining}

        \addplot[mark=*, thick, color=cyan, mark options={solid, fill=cyan}] 
            coordinates {(0,0.642) (10,0.630) (20,0.615) (30,0.599) (40,0.578) (50,0.556) (60,0.527)};
        
        \addplot[mark=square*, thick, color=cyan, mark options={solid, fill=cyan}] 
            coordinates {(0,0.948) (10,0.945) (20,0.939) (30,0.936) (40,0.936) (50,0.938) (60,0.937)};
        
        \addplot[mark=square*, thick, color=orange, mark options={solid, fill=orange}] 
            coordinates {(0,0.948) (10,0.951) (20,0.955) (30,0.961) (40,0.961) (50,0.961) (60,0.973)};
        
        \addplot[mark=*, thick, color=orange, mark options={solid, fill=orange}] 
            coordinates {(0,0.643) (10,0.647) (20,0.655) (30,0.656) (40,0.649) (50,0.649) (60,0.739)};

        \addplot[mark=triangle*, thick, color=cyan, mark options={solid, fill=cyan}] 
            coordinates {(0,1) (10,0.9) (20,0.8) (30,0.7) (40,0.6) (50,0.5) (60,0.4)};
    
        \addplot[mark=triangle*, thick, color=orange, mark options={solid, fill=orange}] 
            coordinates {(0,1) (10,0.856) (20,0.718) (30,0.597) (40,0.555) (50,0.555) (60,0.293)};
        
        \end{axis}
    \end{tikzpicture}

    \caption{\textbf{Llama 3 70B Performance on EPFL Dataset Across Different Cutoff Percentages.} (with Separated Thresholding and Number of MCQs for the Corresponding Percentiles). \textbf{NC 4x:} Evaluation with no context correct for all 4 bias mitigation evaluations of each question; \textbf{WC 4x:} Evaluation with context correct for all 4 bias mitigation evaluations of each question. \textbf{Questions Remaining:} Fraction of the original number of questions still remaining for the given cutoff threshold. \textbf{Cosine:} Only cosine thresholding applied. \textbf{Rouge-L and Jaccard:} Only Rouge-L and Jaccard thresholding applied.}

    \label{fig:llama_70b_epfl_cutoffs}
\end{figure}
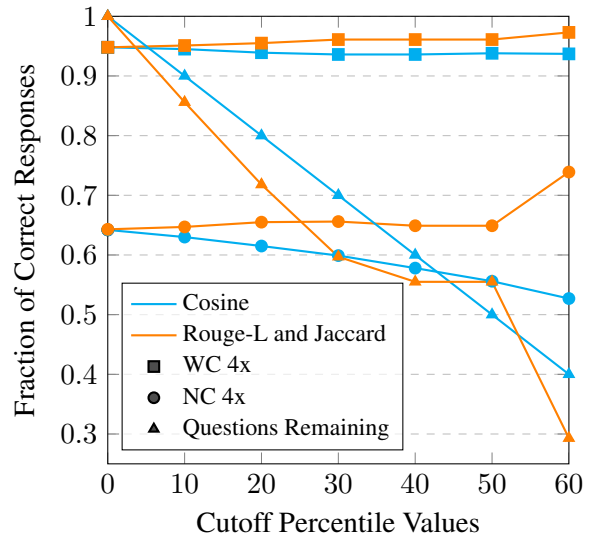

Across five corpora and two models, our pipeline (i) increases MCQ quality via complementary similarity filters, (ii) assigns higher information potential (IP) to specialized/rare collections than to widely available ones, and (iii) yields consistent cross-model patterns. Figure \ref{fig:llama_70b_epfl_cutoffs} illustrates the impact of different thresholding methods, while Figure~\ref{fig:venn_statistics_overall} summarizes correctness with/without context and the resulting IP per dataset/model.

\subsection{Effectiveness of Similarity Thresholding}

\noindent\textbf{Cosine similarity.} Increasing the cutoff reduces performance without context (NC 4x) by about 10\% between 0 and the 50th percentile, while performance with context (WC 4x) stays stable—indicating higher difficulty for no-context answers without hurting context-based performance.\\

\noindent\textbf{ROUGE-L and Jaccard.} In contrast, these filters slightly improve WC 4x accuracy (+2\% up to the 50th percentile) while NC 4x remains stable, suggesting improved question clarity rather than difficulty. However, combined percentile filtering sharply reduces the number of retained questions between the 50th and 60th percentiles.\\

\noindent\textbf{Comparison.} Cosine thresholding mainly increases difficulty, whereas ROUGE-L and Jaccard refine clarity at the cost of coverage.



\subsection{Information Potential Across Corpora}
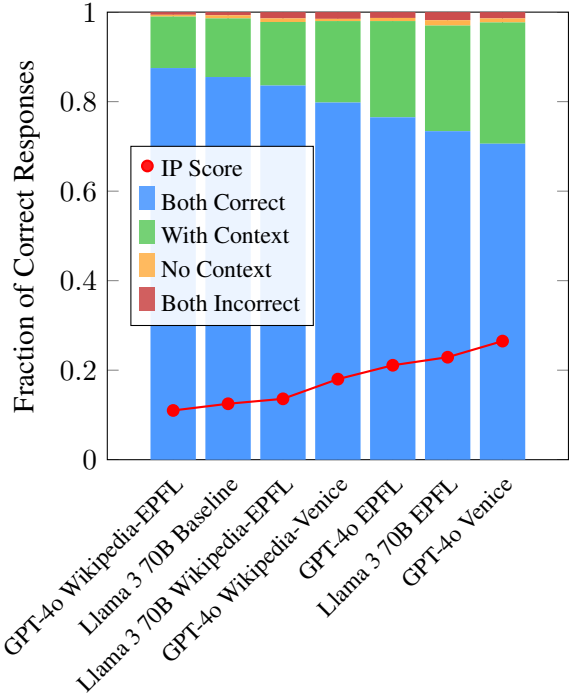
\begin{figure}[h] 
    \centering
    \begin{tikzpicture}
        \pgfplotsset{
            selective show sum on top/.style={
                /pgfplots/scatter/@post marker code/.append code={%
                    \ifnum\coordindex=#1
                       \node[
                       at={(normalized axis cs:%
                           \pgfkeysvalueof{/data point/x},%
                           \pgfkeysvalueof{/data point/y})%
                       },
                       anchor=center,
                       text=black
                       ]
                       {\pgfmathprintnumber{\pgfkeysvalueof{/data point/y}}};
                    \fi
                },
            },selective show sum on top/.default=0
        }
    
        \begin{axis}[
            ybar stacked,
            ymin=0, ymax=1,
            width=0.48\textwidth, height=7.5cm,
            bar width=6mm,
            symbolic x coords={GPT-4o Wikipedia-EPFL,Llama 3 70B Baseline,Llama 3 70B Wikipedia-EPFL,GPT-4o Wikipedia-Venice,GPT-4o EPFL,Llama 3 70B EPFL,GPT-4o Venice},
            xtick=data,
            xticklabel style={rotate=45, anchor=north east, font=\small},
            enlarge x limits=0.2,
            ylabel={Fraction of Correct Responses},
            legend style={font=\small, 
                          fill opacity=0.9,
                          legend cell align=left,
                          mark options={yshift=-5pt}, 
                          fill=white, 
                          text opacity=1,
                          at={(0.05,0.5)},anchor=west}
        ]
    
        \addplot+[
            sharp plot,
            stack plots=false,
            color = red,
            mark = *,
            fill = none,
            thick, 
        ] coordinates {
            (GPT-4o Wikipedia-EPFL, 0.110)
            (Llama 3 70B Baseline, 0.125)
            (Llama 3 70B Wikipedia-EPFL, 0.136)
            (GPT-4o Wikipedia-Venice, 0.180)
            (GPT-4o EPFL, 0.211)
            (Llama 3 70B EPFL, 0.229)
            (GPT-4o Venice, 0.265)
        };
        
        \addplot+[ybar, fill={rgb,1:red,0.3;green,0.6;blue,1}, draw=none, draw opacity=0] plot coordinates {
            (GPT-4o Wikipedia-EPFL,0.875) 
            (Llama 3 70B Baseline,0.855) 
            (Llama 3 70B Wikipedia-EPFL,0.836) 
            (GPT-4o Wikipedia-Venice,0.798) 
            (GPT-4o EPFL,0.765) 
            (Llama 3 70B EPFL,0.734) 
            (GPT-4o Venice, 0.706)};
        
        \addplot+[ybar, fill={rgb,1:red,0.4;green,0.8;blue,0.4}, draw=none, draw opacity=0] plot coordinates {
            (GPT-4o Wikipedia-EPFL,0.115) 
            (Llama 3 70B Baseline,0.131) 
            (Llama 3 70B Wikipedia-EPFL,0.142) 
            (GPT-4o Wikipedia-Venice, 0.182) 
            (GPT-4o EPFL,0.215) 
            (Llama 3 70B EPFL,0.236) 
            (GPT-4o Venice, 0.271)};
        
        \addplot+[ybar, fill={rgb,1:red,1;green,0.7;blue,0.3}, draw=none, draw opacity=0, every node near coord/.append style={yshift=-0.7mm}] plot coordinates {
            (GPT-4o Wikipedia-EPFL,0.004) 
            (Llama 3 70B Baseline,0.007) 
            (Llama 3 70B Wikipedia-EPFL,0.008) 
            (GPT-4o Wikipedia-Venice, 0.005) 
            (GPT-4o EPFL,0.007) 
            (Llama 3 70B EPFL,0.012) 
            (GPT-4o Venice, 0.009)};
        
        \addplot+[ybar, fill={rgb,1:red,0.8;green,0.3;blue,0.3}, draw=none, draw opacity=0, every node near coord/.append style={yshift=0.7mm}] plot coordinates {
            (GPT-4o Wikipedia-EPFL,0.006) 
            (Llama 3 70B Baseline,0.007) 
            (Llama 3 70B Wikipedia-EPFL,0.014) 
            (GPT-4o Wikipedia-Venice,0.015)
            (GPT-4o EPFL,0.013) 
            (Llama 3 70B EPFL,0.018)
            (GPT-4o Venice, 0.014)
        };
        
        \legend{IP Score, Both Correct, With Context, No Context, Both Incorrect} 
        
    \end{axis}
    \end{tikzpicture}
    \caption{\textbf{Information Potential Analysis Across Datasets and Models.}
    Stacked barplot showing correct response overlap between context-free and context-provided conditions. Overlaying the barplot is a line plot showing the Information Potential (IP) scores. Higher IP scores indicate greater novel information content, with PhD manuscripts (EPFL) showing consistently higher IP (0.211-0.229) compared to Wikipedia (0.110-0.136) and synthetic baseline (0.125). Both open and closed-source models exhibit similar patterns despite architectural differences.}
    \label{fig:venn_statistics_overall}
\end{figure}

\noindent\textbf{Limited-Access Knowledge Collections (IP: 0.211–0.265).}
EPFL PhD manuscripts and Venetian historical records yield the highest information potential, driven by large gaps between with-context (WC) and no-context (NC) performance. WC accuracy is consistently $>97\%$ while NC sits around $70$–$73\%$, producing IP $\approx 0.211$–$0.265$. This suggests these collections contain knowledge that is hard to retrieve from model priors and thus valuable for integration (Fig.~\ref{fig:venn_statistics_overall}).\\

\noindent\textbf{Wikipedia Datasets (IP: 0.110–0.180).}
Both EPFL-related and Venice-related Wikipedia subsets show moderate IP with smaller WC–NC gaps (WC $\sim 97\%$ vs.\ NC $\sim 82$–$84\%$). The attenuated gain indicates substantial overlap with pretraining data or readily retrievable priors, aligning with their broad web availability (Fig.~\ref{fig:venn_statistics_overall}).\\

\noindent\textbf{Synthetic Baseline (IP: 0.125).}
Model-generated text forms a lower bound: WC remains near ceiling ($\sim 99\%$) and NC is high ($\sim 85$–$86\%$), yielding a small but nonzero IP $\approx 0.125$. Together with the 4$\times$ rotation consistency checks (Fig.~\ref{fig:venn_statistics_4x_eval}), these trends support IP as a stable proxy for identifying corpora with genuinely novel, harder-to-retrieve information.

\subsection{Comparison between open and closed source model}
Both Llama~3~70B and GPT-4o show similar dataset rankings for information potential: EPFL manuscripts consistently score about twice as high as Wikipedia. GPT-4o yields slightly lower IP than Llama~70B, suggesting stronger baseline knowledge retention, but both models reach near-perfect accuracy ($>97\%$) with context, confirming robust comprehension when relevant text is provided.\\

A broader survey of models could clarify how architecture and scale shape IP, but here our focus is on validating it as a general metric across different model families.

\section{Conclusion}
This work introduced a lightweight pipeline to estimate the information potential (IP) of text collections for LLMs without retraining or fine-tuning. Our contributions are:
\begin{enumerate}[nolistsep]
\item \textbf{Evaluation pipeline:} Automated MCQ generation, filtering, and position-debiased testing provide a scalable measure of collection value.
\item \textbf{Empirical validation:} Results align with expectations---IP rises from synthetic baseline (0.125) to Wikipedia (0.136–0.180) to PhD manuscripts (0.229) and Venetian records (0.265). Larger models (GPT-4o) show lower IP than smaller ones (Llama 3 70B), consistent with stronger prior knowledge.
\end{enumerate}

By highlighting where LLMs benefit most from external context, our method offers a practical screening tool for prioritizing digitization and integration of specialized corpora, from recent academic research to rare historical sources.

\section{Limitations}
\noindent\textbf{Practical Considerations. }Our pipeline generates MCQs before filtering; integrating quality checks directly into generation could improve efficiency. A further challenge is verifying whether corpora truly lie outside LLM training data, given opaque and ever-growing pretraining sets. Finally, our method identifies promising collections but does not address how to integrate them into models without disrupting existing knowledge \citep{moiseev-etal-2022-skill,liu-etal-2024-untangle}.\\

\noindent\textbf{Evaluation Scope. }This study omits human annotation and pretraining experiments to keep the method lightweight. Instead, we rely on safeguards such as similarity filtering, position-debiased testing, and cross-model evaluation. While fine-tuning or human validation could strengthen findings, our results already align with intuitive knowledge gaps (e.g., low IP for synthetic, high IP for specialized corpora).\\

\noindent\textbf{Broader Perspective. }The method requires digitized text but supports sampling-based evaluation—e.g., testing a few chapters before digitizing entire collections—making it useful for resource-constrained projects. More generally, IP highlights areas where models lack or fail to retrieve knowledge, raising broader questions about the line between knowledge possession and competence, and offering a practical complement to theoretical analyses of internal representations.

\bibliography{acl_latex}

\newpage
\appendix
\onecolumn
\section{Appendix}
\label{sec:appendix}

\subsection{Supplementary Methods}
\subsubsection{LLM Prompts}

\begin{table*}[ht]
    \centering
    \footnotesize
    \setlength{\tabcolsep}{5pt} 
    \renewcommand{\arraystretch}{1.5}
    \begin{tabular}{|p{0.2\textwidth}|p{0.7\textwidth}|} 
    \hline
    \normalsize \textbf{Task} & \normalsize \textbf{LLM Prompt} \\
    \hline
    
    \textbf{Multiple-Choice Question Generation} & 
    From the following piece of a scientific PhD manuscript: \newline
    \texttt{'TEXT\_HERE'} \newline
    Design a multiple-choice question with four answers: 'A', 'B', 'C', 'D'. Please provide the correct answer. The question needs to be difficult, but answers should not be ambiguous. Start the question with ['QUESTION'] and the answers with 'A', 'B', 'C', 'D'. Be concise! \newline
    Please generate a total of 10 MCQs. Avoid references to the manuscript itself (e.g., do not use phrases like 'according to the text,' 'as stated in the manuscript,' or 'based on the passage' etc.). Use the following format: '\texttt{[QUESTION] <question>} \newline
    \texttt{A) <option A>} \newline
    \texttt{B) <option B>} \newline
    \texttt{C) <option C>} \newline
    \texttt{D) <option D>} \newline
    \texttt{Correct answer: <correct answer letter>) <correct answer>}' \\
    \hline
    
    \textbf{Multiple Choice Question Answer Generation} & 
    For the following multiple choice question: \newline
    \texttt{'QUESTION\_TEXT\_HERE'} \newline
    Please write which answer option (A, B, C, or D) is the correct one. Answer in the following format: '\texttt{Correct answer: <answer letter>.}' \\
    \hline
    
    \textbf{Context-Based Multiple Choice Question Answer Generation} & 
    Using the information of the following passage: \newline
    \texttt{'PASSAGE\_TEXT\_HERE'} \newline
    Answer the following multiple-choice question: \newline
    \texttt{'QUESTION\_TEXT\_HERE'} \newline
    Please write which answer option (A, B, C, or D) is the correct one. Answer in the following format: '\texttt{Correct answer: <answer letter>.}' \\
    \hline
    
    \textbf{Baseline Subtopic List Generation} & 
    For the following topic: \newline
    \texttt{'TOPIC\_HERE'} \newline
    Please generate a list of 5 subtopics that could be used to create a comprehensive PhD manuscript about this topic. List them in order and number them in the following format: '\texttt{1) <write subtopic 1 here>} \newline
    \texttt{2) <write subtopic 2 here>} \newline
    \texttt{3) <write subtopic 3 here>} \newline
    \texttt{4) <write subtopic 4 here>} \newline
    \texttt{5) <write subtopic 5 here>}\newline
    \texttt{<end>}' \\
    \hline
    
    \textbf{Baseline Chapter Generation} & 
    For a scientific manuscript with the following title: \newline
    \texttt{'MANUSCRIPT\_TITLE\_HERE'} \newline
    Please generate a comprehensive chapter that covers the following subtopic: \texttt{'SUBTOPIC\_HERE'}. Aim for around 600 words, include facts and numbers, and focus solely on substantial content. Omit any introductory or closing remarks and just output the content that this chapter would have. \\
    \hline
    \end{tabular}
    \caption{LLM Task Prompt Templates}
    \label{tab:prompts}
\end{table*}

\newpage
\subsection{Supplementary Results}
In the following we present additional results for the baseline, Wikipedia-EPFL, and EPFL datasets.

\subsubsection{Positional Bias}
For all three generated MCQ datasets, GPT-4o shows a strong tendency to place the correct answer in the answer options B and C over A and D in around 80\% of the time. This tendency may come from training biases where datasets exhibited a similar distribution in MCQ formats.

When ignored, this positional bias may skew a model's performance during evaluation. To counteract this effect, this project employed the rotation of the correct answer position and evaluated each question four times independently. This ensures a balanced distribution of the correct answer among the four positions and reduces the risk of skewing the evaluation statistics with the positional MCQ generation bias. After the positional bias mitagation strategy is applied, the correct answer is distrubuted evenly, appearing in each option 25\% of the time. During evaluation, the models also show some levels of positional bias, however on a lower scale than during the MCQ generation.

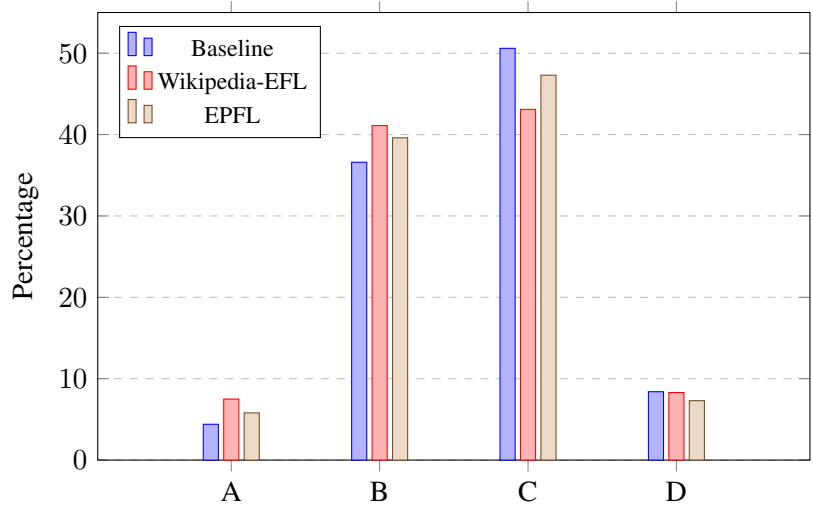
\begin{figure}[h] 
    \centering
    \begin{tikzpicture}
        \begin{axis}[
            width=11cm,
            height=7.5cm,
            ybar,
            bar width=0.2cm,
            symbolic x coords={A,B,C,D},
            xtick=data,
            ymin=0, ymax=55,
            ylabel={Percentage},
            legend pos=north west,
            nodes near coords,
            enlarge x limits=0.3,
            ymajorgrids=true,
            grid style=dashed,
            nodes near coords={},
            legend style={font=\small}
        ]
            \addplot coordinates {(A,4.4) (B,36.6) (C,50.6) (D,8.4)};
            \addplot coordinates {(A,7.5) (B,41.1) (C,43.1) (D,8.3)};
            \addplot coordinates {(A,5.8) (B,39.6) (C,47.3) (D,7.3)};
            
            \legend{Baseline, Wikipedia-EFL, EPFL}
        \end{axis}
    \end{tikzpicture}
    \caption{Distribution of the Correct Answer Among the Answer Options for the MCQ Dataset Generated with GPT-4o Before Positional Bias Mitigation.}
    \label{fig:positional_bias_gpt_mcq_generation}
\end{figure}

\begin{figure}[h] 
    \centering
    \begin{tikzpicture}
        \begin{axis}[
            width=11cm,
            height=7.5cm,
            ybar,
            bar width=0.2cm,
            symbolic x coords={A,B,C,D},
            xtick=data,
            ymin=0, ymax=55,
            ylabel={Percentage},
            legend pos=north west,
            enlarge x limits=0.3,
            ymajorgrids=true,
            grid style=dashed,
            nodes near coords={},
            legend style={font=\small}
        ]
            \addplot coordinates {(A,22.78) (B,26.23) (C,27.43) (D,23.55)};
            \addplot coordinates {(A,24.83) (B,25.02) (C,25.30) (D,24.85)};
            \addplot coordinates {(A,24.08) (B,25.39) (C,26.14) (D,24.38)};
            \addplot coordinates {(A,24.87) (B,24.95) (C,25.20) (D,24.98)};
            
            \legend{EPFL - No Context, EPFL - With Context, Wikipedia-EPFL - No Context, Wikipedia-EPFL - With Context}
        \end{axis}
    \end{tikzpicture}
        
    \caption{Distribution of Correct Answer Letter Prediction for EPFL and Wikipedia MCQ Datasets Evaluated on GPT-4o After Positional Bias Mitigation.}
    \label{fig:positional_bias_mcq_eval_combined}
\end{figure}
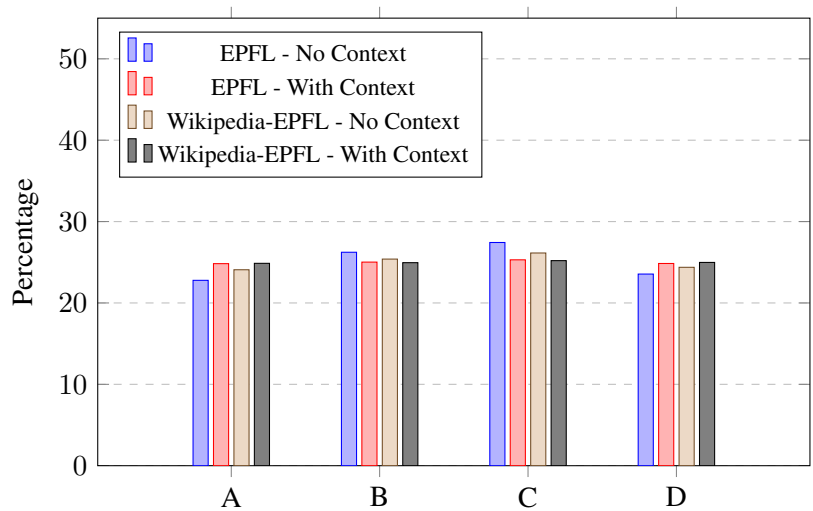

\begin{figure}[h] 
    \centering
    \begin{tikzpicture}
        \begin{axis}[
            width=11cm,
            height=7.5cm,
            ybar,
            bar width=0.2cm,
            symbolic x coords={A,B,C,D},
            xtick=data,
            ymin=0, ymax=55,
            ylabel={Percentage},
            legend pos=north west,
            enlarge x limits=0.3,
            ymajorgrids=true,
            grid style=dashed,
            nodes near coords={},
            legend style={font=\small}
        ]
            \addplot coordinates {(A,25.58) (B,25.48) (C,25.6) (D,23.35)};
            \addplot coordinates {(A,25.14) (B,24.84) (C,25.12) (D,24.9)};
            \addplot coordinates {(A,25.74) (B,24.93) (C,25.31) (D,24.03)};
            \addplot coordinates {(A,24.90) (B,24.98) (C,25.13) (D,24.98)};
            \addplot coordinates {(A,24.71) (B,25.58) (C,25.06) (D,24.64)};
            \addplot coordinates {(A,24.96) (B,25.12) (C,25.04) (D,24.89)};
                
            \legend{EPFL - No Context, EPFL - With Context, Wikipedia-EPFL - No Context, Wikipedia-EPFL - With Context, Baseline - No Context, Baseline - With Context}
        \end{axis}
    \end{tikzpicture}
        
    \caption{Distribution of Correct Answer Letter Prediction for EPFL, Wikipedia, and Baseline MCQ Datasets Evaluated on Llama 3 70B After Positional Bias Mitigation.}
    \label{fig:positional_bias_mcq_eval_llama70b_combined}
\end{figure}
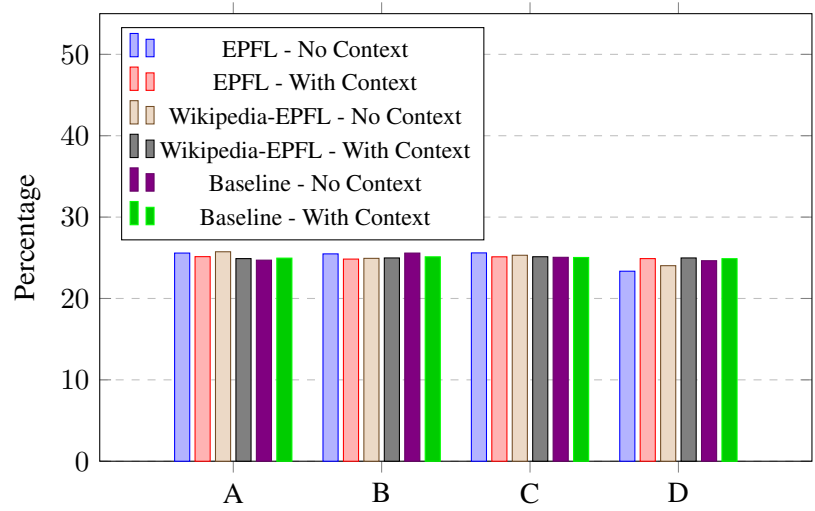

\newpage
\ \newpage
\subsubsection{Information Potential}
\begin{figure*}[h]
    \centering
    \begin{subfigure}[b]{0.31\textwidth}
        \centering
        \includegraphics[width=\textwidth]{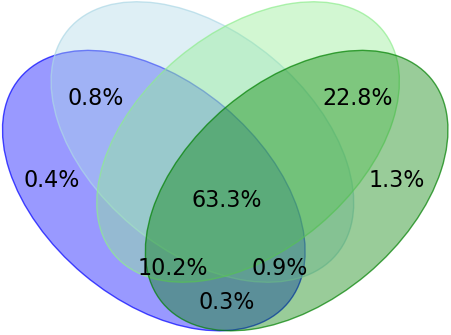}
        \caption{Llama 3 70B EPFL\\IP: 0.229}
        \label{fig:epfl_eval_mcqs}
    \end{subfigure}
    \begin{subfigure}[b]{0.31\textwidth}
        \centering
        \includegraphics[width=\textwidth]{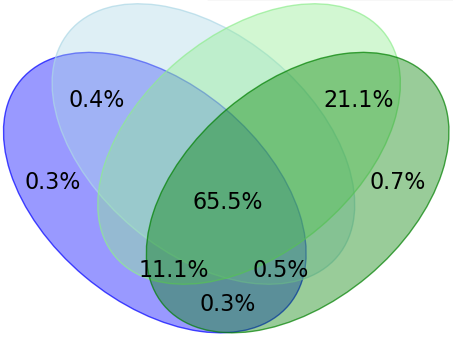}
        \caption{GPT-4o EPFL\\IP: 0.211}
        \label{fig:epfl_eval_books}
    \end{subfigure}
    \begin{subfigure}[b]{0.31\textwidth}
        \centering
        \includegraphics[width=\textwidth]{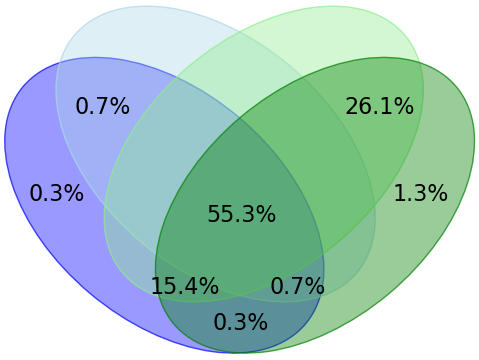}
        \caption{GPT-4o Venice\\IP: 0.265}
        \label{fig:epfl_eval_books}
    \end{subfigure}

    \vspace{0.1cm}

    \begin{subfigure}[b]{0.31\textwidth}
        \centering
        \includegraphics[width=\textwidth]{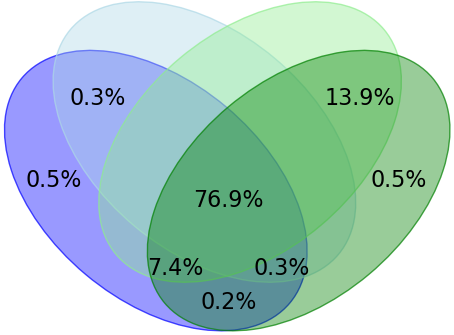}
        \caption{Llama 3 70B Wikipedia-EPFL\\IP: 0.136}
        \label{fig:epfl_eval_wikipedia_2}
    \end{subfigure}
    \begin{subfigure}[b]{0.31\textwidth}
        \centering
        \includegraphics[width=\textwidth]{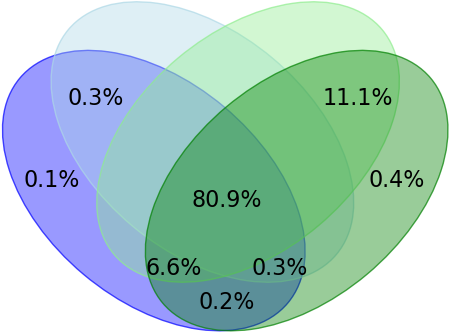}
        \caption{GPT-4o Wikipedia-EPFL\\IP: 0.110}
        \label{fig:epfl_eval_wikipedia}
    \end{subfigure}
    \begin{subfigure}[b]{0.31\textwidth}
        \centering
        \includegraphics[width=\textwidth]{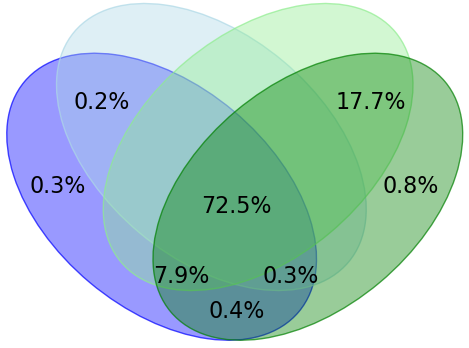}
        \caption{GPT-4o Wikipedia-Venice\\IP: 0.180}
        \label{fig:epfl_eval_books}
    \end{subfigure}

    \vspace{0.1cm}

    \begin{subfigure}[b]{0.31\textwidth}
        \centering
        \includegraphics[width=\textwidth]{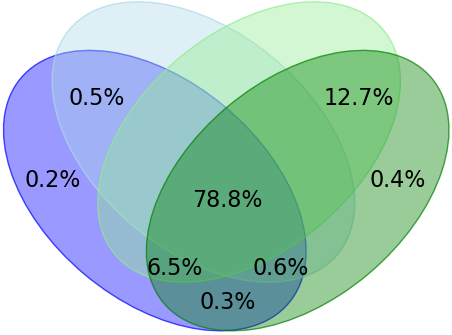}
        \caption{Llama 3 70B Baseline\\IP: 0.125}
        \label{fig:epfl_eval_baseline}
    \end{subfigure}
    \begin{subfigure}[b]{0.31\textwidth}
        \centering
        \includegraphics[width=\textwidth]{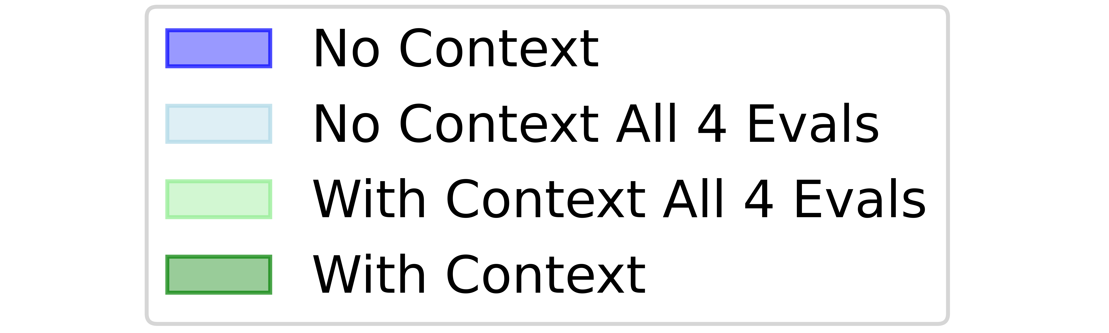}
        \caption{Legend}
        \label{fig:4x_eval_legend}
    \end{subfigure}

    \caption{Statistics of Model Performances Including whether Model was Correct on all Four Evaluations of Each MCQ
    \\Venn diagrams of the number of correctly answered questions when answered with or without context.
    \\ \textbf{IP}: Information Potential computed with Equation \ref{eq:knowledge_gain_formula}}
    \label{fig:venn_statistics_4x_eval}
\end{figure*}

\begin{figure}[h] 
    \centering

    \begin{tikzpicture}
        \begin{axis}[
            width=0.7\textwidth, height=9cm,
            xlabel={Cutoff Percentile Values},
            ylabel={Fraction of Correct Responses},
            xmin=0, xmax=60,
            ymin=0, ymax=1,
            xtick={0,10,20,30,40,50,60},
            ytick={0.2,0.4,0.6,0.8,1},
            legend pos=south west,
            ymajorgrids=true,
            grid style=dashed,
            legend style={font=\small, 
                          fill opacity=0.7,
                          legend cell align=left,
                          fill=white, 
                          text opacity=1,
                          at={(0.05,0.4)},anchor=west}
        ]

        \addplot[thick, color=cyan] 
        coordinates {(0, 0)};
        \addlegendentry{Baseline}

        \addplot[thick, color=violet] 
        coordinates {(0, 0)};
        \addlegendentry{Wikipedia-EPFL}

        \addplot[thick, color=orange] 
        coordinates {(0, 0)};
        \addlegendentry{EPFL}

        \addplot[mark=triangle*, only marks, thick, color=black, mark options={solid, fill=black}] 
            coordinates {(0,0.976)}; 
        \addlegendentry{WC 4x}
        
        \addplot[mark=square*, only marks, thick, color=black, mark options={solid, fill=black}] 
            coordinates {(0,0.796)}; 
        \addlegendentry{NC 4x}
        
        \addplot[mark=*, only marks, thick, color=black, mark options={solid, fill=black}] 
            coordinates {(0,0.125)}; 
        \addlegendentry{IP}

        \addplot[mark=triangle*, thick, color=cyan, mark options={solid, fill=cyan}] 
            coordinates {(0,0.976) (10,0.975) (20,0.977) (30,0.975) (40,0.979) (50,0.983) (60,1)};
        
        \addplot[mark=square*, thick, color=cyan, mark options={solid, fill=cyan}] 
            coordinates {(0,0.796) (10,0.795) (20,0.781) (30,0.766) (40,0.737) (50,0.716) (60,0.854)};
        
        \addplot[mark=*, thick, color=cyan, mark options={solid, fill=cyan}] 
            coordinates {(0,0.125) (10,0.121) (20,0.128) (30,0.138) (40,0.158) (50,0.173) (60,0.089)};
        
        \addplot[mark=triangle*, thick, color=violet, mark options={solid, fill=violet}] 
            coordinates {(0,0.969) (10,0.970) (20,0.971) (30,0.971) (40,0.971) (50,0.966) (60,0.988)};
        
        \addplot[mark=square*, thick, color=violet, mark options={solid, fill=violet}] 
            coordinates {(0,0.770) (10,0.757) (20,0.747) (30,0.741) (40,0.730) (50,0.718) (60,0.718)};
        
        \addplot[mark=*, thick, color=violet, mark options={solid, fill=violet}] 
            coordinates {(0,0.134) (10,0.143) (20,0.148) (30,0.158) (40,0.170) (50,0.173) (60,0.198)};
        
        \addplot[mark=triangle*, thick, color=orange, mark options={solid, fill=orange}] 
            coordinates {(0,0.914) (10,0.924) (20,0.916) (30,0.916) (40,0.916) (50,0.912) (60,0.963)};
        
        \addplot[mark=square*, thick, color=orange, mark options={solid, fill=orange}] 
            coordinates {(0,0.619) (10,0.617) (20,0.611) (30,0.609) (40,0.590) (50,0.565) (60,0.574)};
        
        \addplot[mark=*, thick, color=orange, mark options={solid, fill=orange}] 
            coordinates {(0,0.205) (10,0.211) (20,0.210) (30,0.212) (40,0.226) (50,0.239) (60,0.247)};
        
        \end{axis}
    \end{tikzpicture}
    \caption{Llama 3 70B Performances Across Different Cutoff Percentiles with Joint Cosine and Jaccard-ROUGE-l Metric Thresholding and Number of MCQs for the Corresponding Percentiles.\\
    \textbf{NC 4x:} Evaluation with no context correct for all 4 bias mitigation evaluations of each question.\\
    \textbf{WC 4x:} Evaluation with context correct for all 4 bias mitigation evaluations of each question.\\
    \textbf{IP Score}: The Information Potential computed with Equation \ref{eq:knowledge_gain_formula}
    }
    \label{fig:llama_70b_cutoffs}

\end{figure}
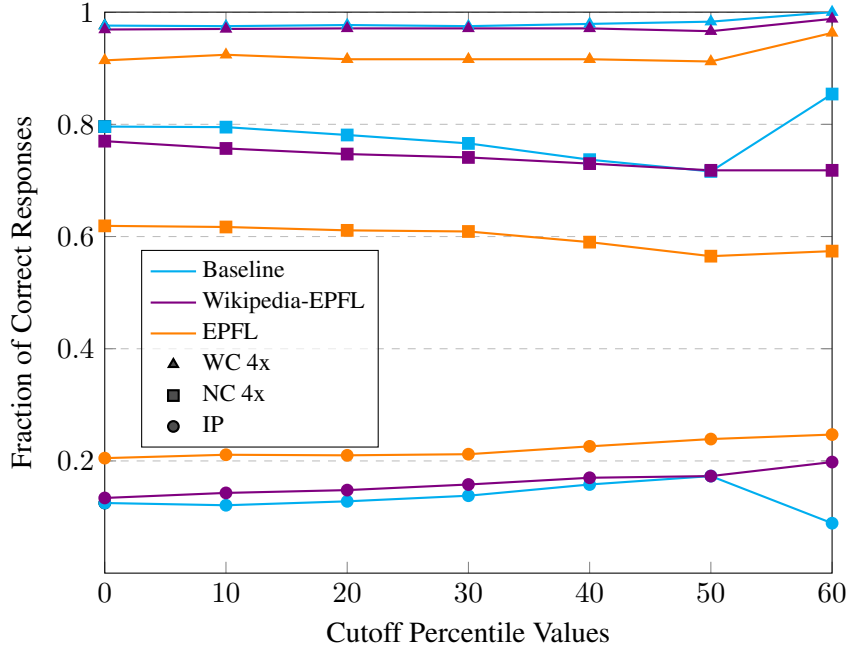

Figure \ref{fig:venn_statistics_4x_eval} shows the models' performance with an emphasis on the positional bias mitigation strategy. As every MCQ is evaluated four times, this allows the analysis of a model's consistency. It is clearly visible that the models become less consistent without context in datasets with higher information potential while this trend is  less pronounced with context.
Figure \ref{fig:llama_70b_cutoffs} shows the performance of Llama 3 70B along with the information potential across the datasets and cutoff percentiles. 

\newpage

\subsubsection{Qualitative Analysis of High-Value Information}

\begin{table*}[h]
    \centering
    \setlength{\tabcolsep}{5pt} 
    \renewcommand{\arraystretch}{1.5}
    \footnotesize
    \begin{tabular}{|p{7cm}|p{7cm}|}
        \hline
        \normalsize \textbf{MCQ} (correct answer in italics) & \normalsize \textbf{Relevant Context Passages}\\ 
        \hline
        \textbf{What is a common feature of 'interference errors' and 'reduction of intentionality errors'?} \newline
          \textit{A) Both involve replacing the correct subject with another similar one.} \newline
          B) Both typically result from incorrect or incomplete mental models. \newline
          C) Both involve action reversals. \newline
          D) Both require complex decision-making at the knowledge-based level.
          &
          [...] Interference errors. These errors occur when people multi-task, i.e., when multiple sequences of action are active at the same time, which can result in a person combining them. [...] this error and the ones for reduced intentionality and perceptual confusions can be modeled by replacing the correct subject with another one. [...] Reversals [...] cause a person to undo a previously performed action
          \\
        \hline
        \textbf{Which error describes memorizing an action without proceeding to its next logical step?} \newline
          \textit{A) Repetition.} \newline
          B) Omission. \newline
          C) Timing error. \newline
          D) Sequence error. 
          &
          [...] Repetitions. These errors cause a person to misjudge the progress of a sequence of actions, making them perform an action already carried on. [...] Omission, when a user skips the current action and execution continues with its successor, e.g., jumping from action i to action i + 1 [...] Timing errors, when users interact with a system at the wrong time, e.g., too early or too late [...] Sequence errors, when users execute an action out of order [...]
          \\
        \hline

        \textbf{How are articulatory features (AF) different from phone posterior features in terms of prediction?} \newline
          A) AFs rely on spectral analysis. \newline
          B) AFs use a frame-to-phoneme alignment. \newline
          \textit{C) AFs map phonemes to articulatory features.} \newline
          D) AFs predict phonemes directly. 
          &
          [...] There are different ways to represent phonemes as articulatory features, e.g. as binary features (Chomsky and Halle, 1968) or multi-valued features (Ladefoged, 1993). Similar to phone posterior features, they are trained from a frame-to-phoneme alignment. However, instead of predicting phonemes, a mapping from phones to AF is used as targets of the predictor. [...] AFs are modeled by 18 off-the-shelf recurrent neural networks (RNN) based binary classifiers, i.e. D = 18 × 2. The RNNs take as input log energies of 33-dimensional Mel filterbank energies. [...] Similar to phone posterior features, [AFs] are trained from a frame-to-phoneme alignment [...]
          \\
        \hline
    \end{tabular}
    \caption{\textbf{Examples of technical terminology, unique mentions, and complex relationships in EPFL PhD manuscripts.} MCQs requiring context for correct model responses}
    \label{tab:mcq_analysis}
\end{table*}

The MCQs presented in Table \ref{tab:mcq_analysis} exemplify three key patterns in identifying valuable information:

\begin{enumerate}[noitemsep]
\item \textbf{Technical Terminology:} The question about articulatory features demonstrates the significant performance gap when dealing with specialized terminology. This question, drawn from speech recognition research, requires specific context to understand how articulatory features differ from phone posterior features in their prediction approach. The model's inability to answer correctly without context highlights how technical domains in PhD manuscripts contain specialized knowledge not captured in pre-training. Such terminology-heavy questions serve as reliable indicators of domain-specific knowledge.

\item \textbf{Unique Mentions:} The question regarding "interference errors" exemplifies how PhD manuscripts capture recent research outcomes. The detailed distinction between interference errors and reduction of intentionality errors represents novel academic insights that weren't available during model pre-training. This type of question effectively identifies valuable new knowledge contributions from academic manuscripts. The consistency with which models fail these questions without context, despite their strong general reasoning capabilities, suggests genuine knowledge gaps rather than reasoning limitations.

\item \textbf{Complex Relationships:} The question about repetition errors showcases the importance of precise contextual information in understanding intricate conceptual relationships. The distinction between repetition and other error types (omission, timing, sequence) requires careful understanding of how these concepts interrelate. This category highlights a key challenge in assessing knowledge: the line between pure knowledge recall and reasoning ability becomes blurred when concepts are interconnected in complex ways. The model's performance on such questions suggests that even sophisticated reasoning capabilities cannot compensate for missing foundational knowledge.
\end{enumerate}
Additional examples of these patterns can be found in Table \ref{tab:mcq_analysis_annex}.

\subsubsection{Qualitative Analysis of EPFL MCQ Dataset}


\begin{table*}[h]
    \centering
    \setlength{\tabcolsep}{3pt}  
    \renewcommand{\arraystretch}{1.3}  
    \footnotesize
    \begin{tabular}{|p{0.45\textwidth}|p{0.4\textwidth}|p{0.1\textwidth}|}
        \hline
        \textbf{MCQ} (correct answer in italics) & \textbf{Relevant Context Passages} & \textbf{Category}\\ 
        \hline
        
        \textbf{What is the only possible scheme for Bernstein wave excitation in the TCV tokamak due to its plasma equilibria?} \newline
          \textit{A) O-SX-B scheme.} \newline
          B) FX-B scheme. \newline
          C) EB-B scheme. \newline
          D) SX-O-B scheme.
          &
          \scriptsize [...] As a consequence, the extremely steep density gradients necessary for the FX-B mode conversion cannot be obtained in TCV plasma equilibria and the only possible Bernstein waves excitation scheme in TCV is the O-SX-B double mode conversion. [...]
          & Technical Terminology \\
        \hline

        \textbf{What technological challenge is associated with reducing power consumption in CMOS circuits?} \newline
          \textit{A) Maintaining acceptable dynamic range in the face of digital noise} \newline
          B) Reducing the intrinsic capacitance per unit area. \newline
          C) Ensuring constant voltage swing at all frequencies. \newline
          D) Achieving higher gain at lower supply voltages.
          &
          \scriptsize [...] In downscaled processes with lower supply voltages, the coupling and noise through the substrate is higher, partially because of the limitations of the substrate and well bias [40,41]. Therefore, sometimes noise that is produced by the chip due to the digital blocks may be orders of magnitude above the thermal noise, so to achieve the required dynamic range we require a proportional increase in power. [...]
          & Technical Terminology \\
        \hline
        
        \textbf{In learning molecule representations directly in the sparse code domain, what is the main constraint imposed on sparse codes?} \newline
          \textit{A) They must be linear combinations of deformed molecules.} \newline
          B) They must be nonlinear mixtures of entire signal sets. \newline
          C) They must consist of deactivated elements. \newline
          D) They must strictly adhere to original signal morphology.
          &
          \scriptsize [...] We constrain sparse codes to be linear combinations of a few, possibly deformed, molecules and we design an algorithm that can learn the structure from the codes without transforming them back into the signal domain. [...]
          & Technical Terminology \\
        \hline

        \textbf{Which component significantly contributes to total variance in back-to-back scan-rescan scenarios?} \newline
          A) 2-week-gap variance. \newline
          \textit{B) Scan-rescan variability} \newline
          C) Session-dependent offsets. \newline
          D) Repositioning effects.
          &
          \scriptsize [...] The scan-rescan differences in back-to-back scanning scenario significantly contributed to the total variance and represented a significant proportion of between-subject variance for all of the investigated structures. [...] Both repositioning (R2) and 2-week-gap between a rescan (R3) did not significantly contribute to the total variability compared to back-to-back scans and between-subject variability. [...]
          & Complex Relationship \\
        \hline
        
        \textbf{Which factor most critically affects the measurement noise in an ex-situ detection setup?} \newline
          \textit{A) The frequency at which measurements are taken.} \newline
          B) The remanence of the magnetic core. \newline
          C) The sensitivity of the lock-in amplifier. \newline
          D) The microbead placement precision.
          &
          \scriptsize [...] However, this increases the measurement noise, as the measurement is carried out in the 1/f noise frequency range. [...]
          & Complex Relationship \\
        \hline
        
        \textbf{What condition allows the bond in the RMIB model to be unbreakable under compressive deformation?} \newline
          \textit{A) High hydrostatic compressive stress.} \newline
          B) High thermal conductivity. \newline
          C) Low volumetric strain. \newline
          D) Low thermal resistance.
          &
          \scriptsize [...] It is known that the hydrostatic compressive strength is infinite for most materials, which means the bond in RMIB model for these cases cannot be broken under compressive deformation. [...]
          & Complex Relationship \\
        \hline

        \textbf{What is the approach used by RouLette to manage materialization overhead in symmetric joins?} \newline
          \textit{A) Symmetric join pruning of tuples forming outputs.} \newline
          B) Incremental materialization of queried tuples. \newline
          C) Partial materialization of all relations. \newline
          D) Deferred materialization until query execution.
          &
          \scriptsize [...] Symmetric joins require that all relations be materialized and hence incur materialization overhead. To reduce the overhead, RouLette materializes only tuples that can form output tuples for their query-set. We call this symmetric join pruning [...]
          & Unique Mentions \\
        \hline
        
        \textbf{What primary limitation affects the clinical success of MPCs in bone healing?} \newline
          \textit{A) Limited number of available endogenous MPCs.} \newline
          B) Extensive proliferation in vitro. \newline
          C) Over-differentiation into non-mesenchymal lineages. \newline
          D) High heterogeneity in cell populations.
          &
          \scriptsize [...] its clinical outcome was rather disappointing 4 . One of reasons for this seems to be the limiting number of available endogenous mesenchymal progenitor cells (MPCs) that can give rise to bone cells. [...]. Hence, there is a clear clinical need for implants that augment the homing/recruitment of endogenous MPCs to fracture sites [...]
          & Unique Mentions \\
        \hline
        
        \textbf{What leads to the gradual increase in average THC concentration over time during oscillations?} \newline
          \textit{A) Accumulation of carbonates on ceria sites.} \newline
          B) Thermal degradation of the catalyst. \newline
          C) Continuous ceria site activation. \newline
          D) Increasing gas hour space velocity (GHSV).
          &
          \scriptsize [...] The higher average THC concentrations levels with time was caused by the gradual accumulation of carbonates on ceria sites during the periodic oscillations. [...]
          & Unique Mentions \\
        \hline

    \end{tabular}
    \caption{Selection of EPFL Dataset MCQs and their Relevant Context Passages Categorized by Question Type where GPT-4o Required Context to Correctly Answer Consistently}
    \label{tab:mcq_analysis_annex}
\end{table*}

\end{document}